# Is margin preserved after random projection?

Qinfeng Shi                                    JAVEN.SHI@ADELAIDE.EDU.AU
Chunhua Shen                          CHUNHUA.SHEN@ADELAIDE.EDU.AU
Rhys Hill                                        RHYS.HILL@ADELAIDE.EDU.AU
Anton van den Hengel        ANTON.VANDENHENGEL@ADELAIDE.EDU.AU

Australian Centre for Visual Technologies, The University of Adelaide

## Abstract

Random projections have been applied in many machine learning algorithms. However, whether margin is preserved after random projection is non-trivial and not well studied. In this paper we analyse margin distortion after random projection, and give the conditions of margin preservation for binary classification problems. We also extend our analysis to margin for multiclass problems, and provide theoretical bounds on multiclass margin on the projected data.

## 1. Introduction

The margin separating classes of data is a key concept in many existing classification algorithms including support vector machines (SVMs) (Cortes & Vapnik, 1995; Crammer & Singer, 2001) and Boosting (Schapire & Freund, 1998). These classifiers are fundamentally involved in identifying and characterising such margins, and are described in terms of the accuracy and generality with which they do so.

Random projections have attracted much attention within a range of fields including signal processing (Donoho, 2006; Baraniuk et al., 2007), and clustering (Schulman, 2000), largely due to the fact that distances are preserved under such transformations in certain circumstances (Dasgupta & Gupta, 2002). Random projections have also been applied to classification for a variety of purposes (Balcan et al., 2006; Duarte et al., 2007; Shi et al., 2009a;b; 2010). However, whether margin is preserved has not been well studied.

Our primary contributions here are to establish the conditions under which margins are preserved after random projection, and to show that error free margins are preserved for both binary and *multiclass* problems if these conditions are met. We also demonstrate that our results hold for one-parameter multiclass classification, which explains the approach used in Shi et al. (2009a;b).

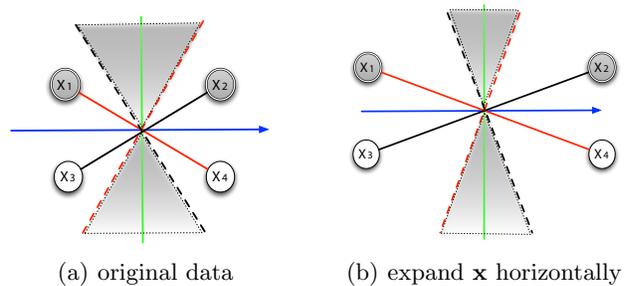

(a) original data       (b) expand **x** horizontally

*Figure 1.* A counter-example illustrating the problem with unnormalised margin preservation. As the data spread further along the horizontal axis it becomes increasingly unlikely (towards probability zero) that a randomly selected projection direction (example shown in blue) will result in a linearly separable projection onto the required subspace (example shown in green). However, during the data spreading, the unnormalised margin of the data before projection is not changed. This means that there are datasets with fixed large positive unnormalised margin, within which the chance of it being linearly separable is zero (corresponding to a negative unnormalised margin).

In this vein we build upon the work of Balcan et al. (2006) which provided a lower bound on the number of dimensions required if a random projection was to have a given probability of maintaining half of the original margin in the data. Although an important step, Balcan et al. (2006) do not solve the problem because the resulting formulation demands infinite many projections in order to guarantee the preservation of an error free margin.

## 2. Motivation and Definitions

A typical definition of margin for a binary classification problem is as follows.

**Definition 1 (Margin)** *The dataset $S = \{(\mathbf{x}_i \in \mathbb{R}^d, y_i \in \{-1, +1\})\}_{i=1}^m$ is said linearly separable by margin $\gamma$ if there exists a unit length $\mathbf{u} \in \mathbb{R}^d$ such that for all $(\mathbf{x}, y) \in S$,*

$$y\langle \mathbf{u}, \mathbf{x} \rangle \geq \gamma.$$

The maximum (among all $\gamma$) smallest (among all data) margin is

$$\gamma^* = \min_{(\mathbf{x}, y) \in S} y\langle \mathbf{u}, \mathbf{x} \rangle. \tag{1}$$

Unfortunately this margin is not preserved after random projection, which we demonstrate by showing a counter-example, depicted in Figure 1. We construct a dataset of 4 data points $(\mathbf{x}_1, y_1), (\mathbf{x}_2, y_2), (\mathbf{x}_3, y_3), (\mathbf{x}_4, y_4)$, where $y_1 = y_2 = +1$ and $y_3 = y_4 = -1$, and $\mathbf{x}_1 = (-1, 1), \mathbf{x}_2 = (1, 1), \mathbf{x}_3 = (-1, -1), \mathbf{x}_3 = (1, -1)$. Let $\mathbf{R} \in \mathbb{R}^{1,2}$ be a random matrix that maps $\mathbf{x}$ onto a 1 dimensional subspace. If the projected data $\mathbf{R}\mathbf{x}$ are to be linearly separable then the subspace onto which they are projected (a line) must lie within the grey area between the two dashed lines in Figure 1(a). If $\mathbf{R}$ is chosen randomly from a uniform distribution then probability that the projected data is linearly separable (by a positive margin) is the angle separating the dashed lines divide by $\pi$. If we expand $\mathbf{x}$ along horizontal line (pushing $\mathbf{x}_1$ and $\mathbf{x}_3$ to the left, $\mathbf{x}_2$ and $\mathbf{x}_4$ to the right while keeping their vertical coordinates unchanged), the grey area angle shrinks as shown in Figure 1(b). In fact, if we push $\mathbf{x}$ to infinitely far away, the angle reduces to 0 while keeping the original margin $\gamma$ unchanged. This means that there exist data with positive margin, such that the chance of the projected data being linearly separable is close to zero.

### 2.1. Error-allowed margin

Balcan et al. (2006) studied the problem of margin preservation under random projection for binary classification. In doing so they derived a formula for the probability that a margin would be decreased by less than half under a particular projection. They provided two margin definitions below, using dataset $S$ as in Definition 1 and data distribution $\mathcal{D}$.

**Definition 2 (Normalised Margin)** *A dataset $S$ is linearly separable by margin $\gamma$ if there exists $\mathbf{u} \in \mathbb{R}^d$, such that for all $(\mathbf{x}, y) \in S$,*

$$y\frac{\langle \mathbf{u}, \mathbf{x} \rangle}{\|\mathbf{u}\|\|\mathbf{x}\|} \geq \gamma.$$

**Definition 3 (Error-allowed Margin)** *A data distribution $\mathcal{D}$ is linearly separable by margin $\gamma$ with error $\rho$, if there exists $\mathbf{u} \in \mathbb{R}^d$, such that*

$$\Pr_{(\mathbf{x}, y) \sim \mathcal{D}} \left( y\frac{\langle \mathbf{u}, \mathbf{x} \rangle}{\|\mathbf{u}\|\|\mathbf{x}\|} < \gamma \right) \leq \rho.$$

Definition 2 describes a normalised version of the more traditional margin, and it is this normalised version which we refer to as the margin henceforth.

Definition 3 describes a margin over a distribution rather than a dataset. Balcan et al. (2006) showed that if the original data has normalised margin $\gamma$ then as long as the number of projections

$$n \geq \frac{c}{\gamma^2} \ln \frac{1}{\rho\delta}, \tag{2}$$

for an appropriate constant $c$, the projected data (now $n$ dimensional) has margin $\gamma/2$ with error $\rho$, with probability at least $1 - \delta$. Definition 3 shows that a positive margin implies $\rho = 0$, which by (2) implies that $n = +\infty$. Thus in order to preserve a positive margin in the projected data one needs infinitely many random projections.

## 3. Margin Distortion and Preservation

In this section, we will establish the conditions for error free margin preservation. Extension to error allowed margin will be briefly discussed at the end of this section.

In order to give an indication that angles and margins might be preserved we first tackle the simpler question of whether the mean is preserved.

**Lemma 4 (Mean preservation)** *For any $\mathbf{w}, \mathbf{x} \in \mathbb{R}^d$, any random Gaussian matrix $\mathbf{R} \in \mathbb{R}^{n,d}$ whose entries $\mathbf{R}(i, j) = \frac{1}{\sqrt{n}} r_{ij}$ where the $r_{ij}s$ are i.i.d. random variables from $\mathcal{N}(0, 1)$, we have*

$$\mathbb{E}(\langle \mathbf{R}\mathbf{w}, \mathbf{R}\mathbf{x} \rangle) = \langle \mathbf{w}, \mathbf{x} \rangle. \tag{3}$$

**Proof of Lemma 4.**

$$\mathbb{E}(\langle \mathbf{R}\mathbf{w}, \mathbf{R}\mathbf{x} \rangle)$$

$$= \frac{1}{n} \mathbb{E}\left[ \sum_{\ell=1}^{n} \left( \sum_{j=1}^{d} r_{\ell j} w_j \sum_{i=1}^{d} r_{\ell i} x_i \right) \right]$$

$$= \frac{1}{n} \sum_{\ell=1}^{n} \left( \sum_{j=1}^{d} \mathbb{E}(r_{\ell j}^2) w_j x_j \right.$$

$$\left. + \sum_{j=1}^{d} \mathbb{E}(r_{\ell j}) w_j \sum_{i \neq j : i=1}^{d} \mathbb{E}(r_{\ell i}) x_i \right).$$

$$= \langle \mathbf{w}, \mathbf{x} \rangle.$$

The above is based solely on the fact that the $\{r_{ij}\}$ are independent with zero mean and unit variance.

The fact that the mean is preserved is a necessary condition for margin preservation, although the following proves that angle and margin are preserved do not depend on it. Due to the 2-stability of the Gaussian distribution, we know that $\sum_{j=1}^{d} r_{\ell j} w_j = \|\mathbf{w}\| z_\ell$ and $\sum_{j=1}^{d} r_{\ell j} x_j = \|\mathbf{x}\| z'_\ell$, where $z_\ell$ and $z'_\ell \sim \mathcal{N}(0,1)$. We thus see that $\langle \mathbf{R}\,\mathbf{w}, \mathbf{R}\,\mathbf{x} \rangle = \frac{1}{n} \|\mathbf{w}\| \|\mathbf{x}\| \sum_{\ell=1}^{n} z_\ell z'_\ell$. There are two possible cases:

- If $\mathbf{w} = \mathbf{x}$, then $\sum_{\ell=1}^{n} z_\ell^2$ has a chi-square distribution with $n$-degrees freedom. Applying chi-square distribution tail bound (Achlioptas, 2003) implies tight bounds on $\Pr\left(\|\mathbf{R}\,\mathbf{x}\|^2 \leq (1-\epsilon)\|\mathbf{x}\|^2\right)$ and $\Pr\left(\|\mathbf{R}\,\mathbf{x}\|^2 \leq (1+\epsilon)\|\mathbf{x}\|^2\right)$.

- If $\mathbf{w} \neq \mathbf{x}$, $\sum_{\ell=1}^{n} z_\ell z'_\ell$ is a sum of product normal distributed variables. One approach might be to compute the variance using Chebyshev's inequality, but the resulting bound would be very loose. Our first main result will show that if the angle between $\mathbf{w}$ and $\mathbf{x}$ is small, the inner product is well preserved.

**Theorem 5 (Angle preservation)** *For any $\mathbf{w}, \mathbf{x} \in \mathbb{R}^d$, any random Gaussian matrix $\mathbf{R} \in \mathbb{R}^{n,d}$ as defined above, for any $\epsilon \in (0,1)$, if $\langle \mathbf{w}, \mathbf{x} \rangle > 0$, then with probability at least*

$$1 - 6\exp\left(-\frac{n}{2}\left(\frac{\epsilon^2}{2} - \frac{\epsilon^3}{3}\right)\right),$$

*the following holds*

$$\frac{(1+\epsilon)}{(1-\epsilon)} \frac{\langle \mathbf{w}, \mathbf{x} \rangle}{\|\mathbf{w}\| \|\mathbf{x}\|} - \frac{2\epsilon}{(1-\epsilon)} \leq \frac{\langle \mathbf{R}\,\mathbf{w}, \mathbf{R}\,\mathbf{x} \rangle}{\|\mathbf{R}\,\mathbf{w}\| \|\mathbf{R}\,\mathbf{x}\|}$$
$$\leq 1 - \frac{\sqrt{(1-\epsilon^2)}}{(1+\epsilon)} + \frac{\epsilon}{(1+\epsilon)} + \frac{(1-\epsilon)}{(1+\epsilon)} \frac{\langle \mathbf{w}, \mathbf{x} \rangle}{\|\mathbf{w}\| \|\mathbf{x}\|}. \quad (4)$$

With angle preservation, one can easily show inner product preservation by multiplying $\|\mathbf{R}\,\mathbf{w}\| \|\mathbf{R}\,\mathbf{x}\|$ in (4) which, however, results in a looser bound.

This theorem is one of our main results, and underpins the analysis to follow. The proof is deferred to Section 4. This theorem shares similar insight as Magen (2007), in which Magen showed that random projections preserve volumes and distances to affine spaces. A similar result on inner product is obtained in Arriaga & Vempala (2006, Corollary 2).

Since acute angles are provably preserved, we are now ready to see whether the margin is also preserved. Proving margin preservation requires proving that there exists a parameter vector $\mathbf{v} \in \mathbb{R}^n$ such that the dataset $S$ after random projection can still be separated by a certain margin. The proof is achieved essentially by showing that $\mathbf{R}\,\mathbf{u}$ is one such parameter vector $\mathbf{v}$. Using angle preservation from Theorem 5 and the union bound yields the following theorem.

**Theorem 6 (Binary preservation)** *Given any random Gaussian matrix $\mathbf{R} \in \mathbb{R}^{n,d}$ as defined above, if the dataset $S = \{(\mathbf{x}_i \in \mathbb{R}^d, y_i \in \{-1,+1\})\}_{i=1}^{m}$ is linearly separable by margin (the normalised margin in Definition 2) $\gamma \in (0,1]$, then for any $\delta, \epsilon \in (0,1)$ and any*

$$n > \frac{12}{(3\epsilon^2 - 2\epsilon^3)} \ln \frac{6m}{\delta},$$

*with probability at least $1-\delta$, the dataset $S' = \{(\mathbf{R}\,\mathbf{x}_i \in \mathbb{R}^n, y_i \in \{-1,+1\})\}_{i=1}^{m}$ is linearly separable by margin $\gamma - \frac{2\epsilon}{(1-\epsilon)}$.*

**Proof of Theorem 6.** By definition, for all $(\mathbf{x}, y) \in S$, there exists $\mathbf{u}$, such that $\frac{\langle \mathbf{u}, y\,\mathbf{x} \rangle}{\|\mathbf{u}\| \|\mathbf{x}\|} \geq \gamma$. Applying Theorem 5 and the union bound, we have

$$\Pr\left(\exists (\mathbf{x}, y) \in S, \frac{\langle \mathbf{R}\,\mathbf{u}, y\,\mathbf{R}\,\mathbf{x} \rangle}{\|\mathbf{R}\,\mathbf{u}\| \|\mathbf{R}\,\mathbf{x}\|} \leq \gamma - \frac{2\epsilon}{(1-\epsilon)}\right)$$
$$\leq 6m\exp\left(-\frac{n}{2}\left(\frac{\epsilon^2}{2} - \frac{\epsilon^3}{3}\right)\right).$$

Let $\delta = 6m\exp\left(-\frac{n}{2}\left(\frac{\epsilon^2}{2} - \frac{\epsilon^3}{3}\right)\right)$, then solving for $n$ gives the required bound on $n$.

Note that in Theorem 6, the lower bound of the margin after random projection can become negative for certain values of $\epsilon$. A negative margin implies that the projected data are not linearly separable. Since it is a lower bound (not a upper bound), the implication in this case is only that the separability of the projected data can no longer be guaranteed with high probability.

When the lower bound is positive, Theorem 6 indicates that margin separability for binary classification is preserved with high probability under random projection. This may seem at odds with our initial counter-example, but the difference lies in the distinction between Definition 1 and Definition 2. In the counter-example, pushing $\mathbf{x}$s apart reduces the probability of achieving a separable projection (as indicated by the margin defined in Definition 1). However, during this process the margin as defined in Definition 2 is also shrinking to zero, and thus that it is only this diminished margin which need be preserved.

We now consider the multiclass case, and specifically the widely accepted definition of the multiclass margin from Crammer & Singer (2001). The straightforward multiclass extension of the counter-example shown in Figure 1 shows that the multiclass margin is not always preserved under random projection. As in the binary classification case, we thus introduce the normalised multiclass margin.

**Definition 7 (Normalised Multiclass Margin)**
*The multiclass dataset $S = \{(\mathbf{x}_i \in \mathbb{R}^d, y_i \in \mathcal{Y} = \{1, \ldots, L\})\}_{i=1}^m$ is linearly separable by margin $\gamma \in (0, 1]$, if there exists $\{\mathbf{u}_y \in \mathbb{R}^d\}_{y \in \mathcal{Y}}$, such that for all $(\mathbf{x}, y) \in S$*

$$\frac{\langle \mathbf{u}_y, \mathbf{x} \rangle}{\|\mathbf{u}_y\|\|\mathbf{x}\|} - \max_{y' \neq y} \frac{\langle \mathbf{u}_{y'}, \mathbf{x} \rangle}{\|\mathbf{u}_{y'}\|\|\mathbf{x}\|} \geq \gamma.$$

For the purposes of the discussion to follow we will assume the multiclass dataset $S$ to be as in Definition 7.

The smallest maximum $\gamma$ can be found via

$$\gamma^* = \min_{(\mathbf{x}, y) \in S} \left( \frac{\langle \mathbf{u}_y, \mathbf{x} \rangle}{\|\mathbf{u}_y\|\|\mathbf{x}\|} - \max_{y' \neq y} \frac{\langle \mathbf{u}_{y'}, \mathbf{x} \rangle}{\|\mathbf{u}_{y'}\|\|\mathbf{x}\|} \right). \quad (5)$$

As we will see a special case of the above is used in the one-parameter method in Shi et al. (2009a;b) whereby $\mathbf{u}_y = \mathbf{u}_{y'}$ for all $y, y' \in \mathcal{Y}$. Thus the set $\{\mathbf{u}_y \in \mathbb{R}^d\}_{y \in \mathcal{Y}}$ reduces to a single vector.

**Theorem 8 (Multiclass margin preservation)**
*For any multiclass dataset $S$ and any Gaussian random matrix $\mathbf{R}$, if $S$ is linearly separable by margin $\gamma \in (0, 1]$, then for any $\delta, \epsilon \in (0, 1)$ and any*

$$n > \frac{12}{(3\epsilon^2 - 2\epsilon^3)} \ln \frac{6Lm}{\delta},$$

*with probability at least $1 - \delta$, the dataset $S' = \{(\mathbf{R}\mathbf{x}_i \in \mathbb{R}^n, y_i \in \mathcal{Y}\}_{i=1}^m$ is linearly separable by margin $-\frac{(1+3\epsilon)}{(1-\epsilon^2)} + \frac{\sqrt{(1-\epsilon^2)}}{(1+\epsilon)} + \frac{(1+\epsilon)}{(1-\epsilon)}\gamma$.*

**Proof of Theorem 8.** By the margin definition, for all $(\mathbf{x}, y) \in S$

$$\frac{\langle \mathbf{u}_y, \mathbf{x} \rangle}{\|\mathbf{u}_y\|\|\mathbf{x}\|} - \max_{y' \neq y} \frac{\langle \mathbf{u}_{y'}, \mathbf{x} \rangle}{\|\mathbf{u}_{y'}\|\|\mathbf{x}\|} \geq \gamma.$$

Take any single $(\mathbf{x}, y) \in S$, we have by Theorem 5 and union bound that

$$\Pr\left( \frac{\langle \mathbf{R}\mathbf{u}_y, \mathbf{R}\mathbf{x} \rangle}{\|\mathbf{R}\mathbf{u}_y\|\|\mathbf{R}\mathbf{x}\|} \geq 1 - \frac{(1+\epsilon)}{(1-\epsilon)}(1 - \frac{\langle \mathbf{u}_y, \mathbf{x} \rangle}{\|\mathbf{u}_y\|\|\mathbf{x}\|}) \right)$$

$$\geq 1 - 6\exp\left(-\frac{n}{2}(\frac{\epsilon^2}{2} - \frac{\epsilon^3}{3})\right))$$

$$\Pr\left( \forall y' \neq y, \frac{\langle \mathbf{R}\mathbf{u}_{y'}, \mathbf{R}\mathbf{x} \rangle}{\|\mathbf{R}\mathbf{u}_{y'}\|\|\mathbf{R}\mathbf{x}\|} \leq 1 - \frac{\sqrt{(1-\epsilon^2)}}{(1+\epsilon)} \right.$$

$$+ \frac{\epsilon}{(1+\epsilon)} + \frac{(1-\epsilon)}{(1+\epsilon)} \frac{\langle \mathbf{u}_{y'}, \mathbf{x} \rangle}{\|\mathbf{u}_{y'}\|\|\mathbf{x}\|} )$$

$$\geq 1 - 6(L-1)\exp\left(-\frac{n}{2}(\frac{\epsilon^2}{2} - \frac{\epsilon^3}{3})\right).$$

By the union bound, with probability at least $1 - 6Lm\exp\left(-\frac{n}{2}(\frac{\epsilon^2}{2} - \frac{\epsilon^3}{3})\right)$, for all $(\mathbf{x}, y) \in S$, we have

$$\frac{\langle \mathbf{R}\mathbf{u}_y, \mathbf{R}\mathbf{x} \rangle}{\|\mathbf{R}\mathbf{u}_y\|\|\mathbf{R}\mathbf{x}\|} - \max_{y' \neq y} \frac{\langle \mathbf{R}\mathbf{u}_{y'}, \mathbf{R}\mathbf{x} \rangle}{\|\mathbf{R}\mathbf{u}_{y'}\|\|\mathbf{R}\mathbf{x}\|}$$

$$\geq \frac{\sqrt{(1-\epsilon^2)}}{(1+\epsilon)} - \frac{(1+3\epsilon)}{(1-\epsilon^2)} + \frac{(1+\epsilon)}{(1-\epsilon)}\gamma.$$

If we let $\delta = 6Lm\exp\left(-\frac{n}{2}(\frac{\epsilon^2}{2} - \frac{\epsilon^3}{3})\right)$, we get the required lower bound on $n$.

Note the above result can be easily extended to the widely accepted definition of the multiclass unnormalised margin from Crammer & Singer (2001) by bounding the distortion on $\|\mathbf{R}\mathbf{u}_y\|\|\mathbf{R}\mathbf{x}\|$ and $\|\mathbf{R}\mathbf{u}_{y'}\|\|\mathbf{R}\mathbf{x}\|$. However, larger multiclass unnormalised margin does not mean smaller angle (*i.e.* larger normalised margin), thus may not provide tighter preservation.

The definition of the multiclass margin in Definition 7 assumes the existence of a set $\{\mathbf{u}_y \in \mathbb{R}^d\}_{y \in \mathcal{Y}}$. We now consider whether the margin is preserved in the case where there exists only a single parameter vector $\mathbf{u}$.

**Theorem 9 (One-parameter method)** *For any multiclass dataset $S$, and any random Gaussian matrix $\mathbf{R}$, denote by $\mathbf{R}_y \in \mathbb{R}^{n,d}$ the $y$-th sub-matrix of $\mathbf{R}$, that is $\mathbf{R} = [\mathbf{R}_1, \cdots, \mathbf{R}_y, \cdots, \mathbf{R}_L]$. If $S$ is linearly separable by margin $\gamma \in (0, 1]$, then for any $\delta, \epsilon \in (0, 1]$ and any*

$$n > \frac{12}{(3\epsilon^2 - 2\epsilon^3)} \ln \frac{6m(L-1)}{\delta},$$

*there exists a parameter vector $\mathbf{v} \in \mathbb{R}^n$, such that*

$$\Pr\left( \forall (\mathbf{x}, y) \in S, \forall y' \neq y, \right.$$

$$\frac{\langle \mathbf{v}, \mathbf{R}_y \mathbf{x} \rangle - \langle \mathbf{v}, \mathbf{R}_{y'} \mathbf{x} \rangle}{\|\mathbf{v}\|\sqrt{\|\mathbf{R}_y \mathbf{x}\|^2 + \|\mathbf{R}_{y'}\mathbf{x}\|^2}} \geq$$

$$\left. \frac{-2\epsilon}{1-\epsilon} + \frac{1+\epsilon}{\sqrt{2}L(1-\epsilon)}\gamma \right) \geq 1 - \delta. \quad (6)$$

**Proof of Theorem 9.** By the margin definition there exists $\{\mathbf{w}_y \in \mathbb{R}^d\}_{y \in \mathcal{Y}}$, such that for all $(\mathbf{x}, y) \in S$,

$$\frac{\langle \mathbf{w}_y, \mathbf{x} \rangle}{\|\mathbf{w}_y\|\|\mathbf{x}\|} - \frac{\langle \mathbf{w}_{y'}, \mathbf{x} \rangle}{\|\mathbf{w}_{y'}\|\|\mathbf{x}\|} \geq \gamma, \forall y' \neq y.$$

Without loss of generality we assume that $\mathbf{w}_y$ has unit length[1] for all $y$. So now

$$\langle \mathbf{w}_y, \mathbf{x} \rangle - \langle \mathbf{w}_{y'}, \mathbf{x} \rangle \geq \gamma \|\mathbf{x}\|, \forall y' \neq y.$$

---

[1] This can be achieved by normalisation.

This can be rewritten as

$$\langle \mathbf{u}, \mathbf{x} \otimes \mathbf{e}_y \rangle - \langle \mathbf{u}, \mathbf{x} \otimes \mathbf{e}_{y'} \rangle = \langle \mathbf{x} \otimes \mathbf{e}_y - \mathbf{x} \otimes \mathbf{e}_{y'}, \mathbf{u} \rangle \geq \gamma \|\mathbf{x}\|,$$

where $\mathbf{u}$ is a concatenation of all $\mathbf{w}_y$ *i.e.* $\mathbf{u} = [\mathbf{w}_1^\top, \cdots, \mathbf{w}_y^\top, \cdots \mathbf{w}_L^\top]^\top$, $\mathbf{e}_y$ is a vector $\in \mathbb{R}^L$ with 1 at the $y$-th location and zeros in all others, and $\otimes$ is the tensor product. Define $\mathbf{z}_{\mathbf{x},y'} = \mathbf{x} \otimes \mathbf{e}_y - \mathbf{x} \otimes \mathbf{e}_{y'}$.

Applying Theorem 5 to $\mathbf{u}$ and $\mathbf{z}_{\mathbf{x},y'}$, we have for a given $(\mathbf{x}, y)$ and a fixed $y' \neq y$, with probability at least $1 - 6\exp\left(-\frac{n}{2}\left(\frac{\epsilon^2}{2} - \frac{\epsilon^3}{3}\right)\right)$, that the following holds,

$$\frac{\langle \mathbf{R}\,\mathbf{u}, \mathbf{R}\,\mathbf{z}_{\mathbf{x},y'}\rangle}{\|\mathbf{R}\,\mathbf{u}\|\|\mathbf{R}\,\mathbf{z}_{\mathbf{x},y'}\|} \geq 1 - \frac{1+\epsilon}{1-\epsilon}(1 - \frac{\langle \mathbf{x}\otimes\mathbf{e}_y - \mathbf{x}\otimes\mathbf{e}_{y'}, \mathbf{u}\rangle}{\sqrt{2}\|\mathbf{u}\|\|\mathbf{x}\|})$$

$$= 1 - \frac{1+\epsilon}{1-\epsilon} +$$

$$\frac{1+\epsilon}{\sqrt{2}(1-\epsilon)}(\frac{\langle \mathbf{w}_y, \mathbf{x}\rangle}{\|\mathbf{u}\|\|\mathbf{x}\|} - \frac{\langle \mathbf{w}_{y'}, \mathbf{x}\rangle}{\|\mathbf{u}\|\|\mathbf{x}\|})$$

$$\geq 1 - \frac{1+\epsilon}{1-\epsilon} + \frac{1+\epsilon}{\sqrt{2L}(1-\epsilon)}\gamma$$

$$= \frac{-2\epsilon}{1-\epsilon} + \frac{1+\epsilon}{\sqrt{2L}(1-\epsilon)}\gamma$$

By the union bound over $m$ samples and the $L-1$ $y'$s,

$$\Pr\Big(\exists(\mathbf{x}, y) \in S, \exists y' \neq y,$$

$$\frac{\langle \mathbf{R}\,\mathbf{u}, \mathbf{R}\,\mathbf{z}_{\mathbf{x},y'}\rangle}{\|\mathbf{R}\,\mathbf{u}\|\|\mathbf{R}\,\mathbf{z}_{\mathbf{x},y'}\|} < \frac{-2\epsilon}{1-\epsilon} + \frac{1+\epsilon}{\sqrt{2L}(1-\epsilon)}\gamma\Big)$$

$$\leq 6m(L-1)\exp\left(-\frac{n}{2}\left(\frac{\epsilon^2}{2} - \frac{\epsilon^3}{3}\right)\right).$$

Letting $\mathbf{v} = \mathbf{R}\,\mathbf{u}$, we have

$$\langle \mathbf{R}\,\mathbf{u}, \mathbf{R}\,\mathbf{z}_{\mathbf{x},y'}\rangle = \langle \mathbf{v}, \mathbf{R}_y\,\mathbf{x} - \mathbf{R}_{y'}\,\mathbf{x}\rangle.$$

Setting $\delta = 6m(L-1)\exp\left(-\frac{n}{2}\left(\frac{\epsilon^2}{2} - \frac{\epsilon^3}{3}\right)\right)$ gives the required bound on $n$.

Theorem 9 shows the existence of a parameter vector under which the margin is preserved up to an order $O(\gamma/\sqrt{2L})$ term where $n$ increases logarithmically with $L$. The gain is thus that *the memory requirement is independent of the number of classes.*

**Error allowed margin** The error free margin preservation results presented above apply only to linearly separable data. Real data are often not linearly separable, but the result applies none the less to any linearly separable subset of the data. Given that what has been developed is a theoretical result intended to guide the selection of an appropriate projection dimension the fact that it applies to every linearly separable subset of the data is likely to suffice in most cases. Our results can be easily extended to error allowed margin, however, in both binary and multiclass cases, by the addition of a controllable tolerance $\epsilon$. The probability of preserving the error allowed margin can be bounded below by bounding above the chance of the subset of data being not linearly separable (thus error allowed) and the chance of the complement subset being linearly separable under projection.

## 4. Angle Preservation

To prove angle preservation as in Theorem 5, we will use the following tail bound, which also appears in a different form in the simplified proof of the Johnson-Lindenstrauss Lemma in Dasgupta & Gupta (2002).

**Lemma 10 (Tail bound)** *For any $\mathbf{x} \in \mathbb{R}^d$, any random Gaussian matrix $\mathbf{R} \in \mathbb{R}^{n,d}$ as defined above, for any $\epsilon \in (0, 1)$,*

$$\Pr\left((1-\epsilon) \leq \frac{\|\mathbf{R}\,\mathbf{x}\|^2}{\|\mathbf{x}\|^2} \leq (1+\epsilon)\right)$$

$$\geq 1 - 2\exp\left(-\frac{n}{2}\left(\frac{\epsilon^2}{2} - \frac{\epsilon^3}{3}\right)\right).$$

**Proof of Theorem 5.** From Lemma 10 and the union bound, we know that

$$(1-\epsilon) \leq \frac{\|\mathbf{R}\,\mathbf{x}\|^2}{\|\mathbf{x}\|^2} \leq (1+\epsilon), (1-\epsilon) \leq \frac{\|\mathbf{R}\,\mathbf{w}\|^2}{\|\mathbf{w}\|^2} \leq (1+\epsilon) \tag{7}$$

holds with probability at least $1 - 4\exp\left(-\frac{n}{2}\left(\frac{\epsilon^2}{2} - \frac{\epsilon^3}{3}\right)\right)$. When (7) holds, due to the fact that increasing the length of two unit length vectors (*i.e.* from $\frac{\mathbf{R}\,\mathbf{x}}{\|\mathbf{R}\,\mathbf{x}\|}$ and $\frac{\mathbf{R}\,\mathbf{w}}{\|\mathbf{R}\,\mathbf{w}\|}$ to $\frac{\mathbf{R}\,\mathbf{x}}{\sqrt{(1-\epsilon)}\|\mathbf{x}\|}$ and $\frac{\mathbf{R}\,\mathbf{w}}{\sqrt{(1-\epsilon)}\|\mathbf{w}\|}$) increases the norm of their difference[2], we have

$$\|\frac{\mathbf{R}\,\mathbf{x}}{\|\mathbf{R}\,\mathbf{x}\|} - \frac{\mathbf{R}\,\mathbf{w}}{\|\mathbf{R}\,\mathbf{w}\|}\|^2$$

$$\leq \|\frac{\mathbf{R}\,\mathbf{x}}{\sqrt{(1-\epsilon)}\|\mathbf{x}\|} - \frac{\mathbf{R}\,\mathbf{w}}{\sqrt{(1-\epsilon)}\|\mathbf{w}\|}\|^2. \tag{8}$$

We thus see that

$$\|\frac{\mathbf{R}\,\mathbf{x}}{\|\mathbf{x}\|} - \frac{\mathbf{R}\,\mathbf{w}}{\|\mathbf{w}\|}\|^2$$

$$\leq \|\sqrt{(1-\epsilon)}\frac{\mathbf{R}\,\mathbf{x}}{\|\mathbf{R}\,\mathbf{x}\|} - \sqrt{(1+\epsilon)}\frac{\mathbf{R}\,\mathbf{w}}{\|\mathbf{R}\,\mathbf{w}\|}\|^2$$

$$\leq \|\sqrt{(1+\epsilon)}(\frac{\mathbf{R}\,\mathbf{x}}{\|\mathbf{R}\,\mathbf{x}\|} - \frac{\mathbf{R}\,\mathbf{w}}{\|\mathbf{R}\,\mathbf{w}\|})\|^2$$

$$+ (\sqrt{(1+\epsilon)} - \sqrt{(1-\epsilon)})^2. \tag{9}$$

---

[2]Note that the opposite does not hold in general.

The first inequality is due to (7), the second inequality is a property of any acute angle. Applying Lemma 10 to the vector $(\frac{\mathbf{x}}{\|\mathbf{x}\|} - \frac{\mathbf{w}}{\|\mathbf{w}\|})$, we see that

$$
(1-\epsilon)\|\frac{\mathbf{x}}{\|\mathbf{x}\|} - \frac{\mathbf{w}}{\|\mathbf{w}\|}\|^2 \leq \|\frac{\mathbf{R}\,\mathbf{x}}{\|\mathbf{x}\|} - \frac{\mathbf{R}\,\mathbf{w}}{\|\mathbf{w}\|}\|^2
$$
$$
\leq (1+\epsilon)\|\frac{\mathbf{x}}{\|\mathbf{x}\|} - \frac{\mathbf{w}}{\|\mathbf{w}\|}\|^2 \tag{10}
$$

holds with a given probability.

Letting $\beta$ denote the angle between $\mathbf{w}$ and $\mathbf{x}$ we have

$$
\gamma = \frac{\langle \mathbf{w}, \mathbf{x} \rangle}{\|\mathbf{w}\|\|\mathbf{x}\|} = \cos(\beta) = 1 - 2\sin^2(\frac{\beta}{2})
$$
$$
= 1 - \frac{1}{2}\|\frac{\mathbf{x}}{\|\mathbf{x}\|} - \frac{\mathbf{w}}{\|\mathbf{w}\|}\|^2. \tag{11}
$$

Similarly

$$
\frac{\langle \mathbf{R}\,\mathbf{w}, \mathbf{R}\,\mathbf{x} \rangle}{\|\mathbf{R}\,\mathbf{w}\|\|\mathbf{R}\,\mathbf{x}\|} = 1 - \frac{1}{2}\|\frac{\mathbf{R}\,\mathbf{x}}{\|\mathbf{R}\,\mathbf{x}\|} - \frac{\mathbf{R}\,\mathbf{w}}{\|\mathbf{R}\,\mathbf{w}\|}\|^2. \tag{12}
$$

Using (10), (8) and (9) we see that $\|\frac{\mathbf{R}\,\mathbf{x}}{\|\mathbf{R}\,\mathbf{x}\|} - \frac{\mathbf{R}\,\mathbf{w}}{\|\mathbf{R}\,\mathbf{w}\|}\|^2$ is bounded below and above by two terms involving $\|\frac{\mathbf{x}}{\|\mathbf{x}\|} - \frac{\mathbf{w}}{\|\mathbf{w}\|}\|^2$. Plugging (11) and (12) into the two side bounds, we get (4). Here we have applied Lemma 10 to 3 vectors, namely $\mathbf{x}$, $\mathbf{w}$, and $(\frac{\mathbf{x}}{\|\mathbf{x}\|} - \frac{\mathbf{w}}{\|\mathbf{w}\|})$, thus by the union bound, the probability that the above holds is at least $1 - 6\exp\left(-\frac{n}{2}(\frac{\epsilon^2}{2} - \frac{\epsilon^3}{3})\right)$.

For completeness, we show the proof of Lemma 10 below.

**Proof of Lemma 10.** By Lemma 4 and letting $\mathbf{w} = \mathbf{x}$, we have $\mathbb{E}(\|\mathbf{R}\,\mathbf{x}\|^2) = \|\mathbf{x}\|^2$. Due to the 2-stability of the Gaussian distribution, we know $\sum_{j=1}^{d} r_{\ell j} x_j = \|\mathbf{x}\| z_\ell$, where $z_\ell \sim \mathcal{N}(0, 1)$. We thus have $\|\mathbf{R}\,\mathbf{x}\|^2 = \frac{1}{n}\|\mathbf{x}\|^2 \sum_{\ell=1}^{n} z_\ell^2$. Here $\sum_{\ell=1}^{n} z_\ell^2$ is chi-square distributed with $n$-degrees of freedom. Applying the standard tail bound of the chi-square distribution, we have

$$
\Pr\left(\|\mathbf{R}\,\mathbf{x}\|^2 \leq (1-\epsilon)\|\mathbf{x}\|^2\right)
$$
$$
\leq \exp\left(\frac{n}{2}(1 - (1-\epsilon) + \ln(1-\epsilon))\right) \leq \exp\left(-\frac{n}{4}\epsilon^2\right).
$$

Here we used the inequality $\ln(1-\epsilon) \leq -\epsilon - \epsilon^2/2$. Similarly, we have

$$
\Pr\left(\|\mathbf{R}\,\mathbf{x}\|^2 \leq (1+\epsilon)\|\mathbf{x}\|^2\right)
$$
$$
\leq \exp\left(\frac{n}{2}(1 - (1+\epsilon) + \ln(1+\epsilon))\right) \leq \exp\left(-\frac{n}{2}(\frac{\epsilon^2}{2} - \frac{\epsilon^3}{3})\right).
$$

Here we used the inequality $\ln(1+\epsilon) \leq \epsilon - \epsilon^2/2 + \epsilon^3/3$.

## 5. Experiments

The experiments detailed below offer an empirical validation in support of the theoretical analysis above, and a demonstration of its application to SVMs.

### 5.1. Angle and inner product preservation condition

Figure 2 shows the results of simulations whereby we randomly generate two vectors $\mathbf{w}$ and $\mathbf{x} \in \mathbb{R}^d$, $d = 300$. We then generate 2,000 random Gaussian matrices of the form specified in Theorem 5. Each such matrix is used to project the data into $n$ dimensions where $n = \{30, 60, 90, \ldots, 300\}$. We vary $\epsilon \in \{0.1, 0.3\}$ and compute the empirical rejection probability for angle preservation

$$
P_1 = 1 - \Pr\left((1-\epsilon) \leq \frac{\langle \mathbf{R}\,\mathbf{w}, \mathbf{R}\,\mathbf{x} \rangle \|\mathbf{w}\|\|\mathbf{x}\|}{\|\mathbf{R}\,\mathbf{w}\|\|\mathbf{R}\,\mathbf{x}\|\langle \mathbf{w}, \mathbf{x} \rangle} < (1+\epsilon)\right),
$$

and the empirical rejection probability for the inner product preservation

$$
P_2 = 1 - \Pr\left((1-\epsilon) \leq \frac{\langle \mathbf{R}\,\mathbf{w}, \mathbf{R}\,\mathbf{x} \rangle}{\langle \mathbf{w}, \mathbf{x} \rangle} \leq (1+\epsilon)\right).
$$

The simulations cover two cases, first where the vectors $\mathbf{x}$ and $\mathbf{w}$ are separated by an acute angle (*i.e.* $\gamma = \frac{\langle \mathbf{w}, \mathbf{x} \rangle}{\|\mathbf{w}\|\|\mathbf{x}\|} > 0$), and second where the angle is obtuse (*i.e.* $\gamma < 0$).

In the acute angle ($\gamma > 0$) case we generate two pairs of vectors with $\langle \mathbf{w}_1, \mathbf{x}_1 \rangle = 0.827$ and $\langle \mathbf{w}_2, \mathbf{x}_2 \rangle = 0.527$ and plot the empirical rejection probability in Figure 2. In Figure 2(a), we can clearly see that the rejection probability more rapidly approaches zero for $\{\mathbf{w}_1, \mathbf{x}_1\}$ than $\{\mathbf{w}_2, \mathbf{x}_2\}$. This aligns with the theoretical analysis above in that we we would expect a more acute angle to imply a better angle preservation under random projection, and thus that fewer projections would be required to achieve a reasonable separation. It is also expected that the rejection probability decreases as $\epsilon$ and $n$ increase.

Likewise, the inner product is preserved under random projection if the angle is acute as is visible in Figure 2(b). It is interesting to see that for the same $\gamma$, $\epsilon$ and $n$, the empirical rejection probability for the angle preservation is significantly smaller than that for inner product preservation.

In the obtuse angle (*i.e.* $\gamma < 0$) case, we generate two pairs of vectors with $\langle \mathbf{w}_3, \mathbf{x}_3 \rangle = -0.062$ and $\langle \mathbf{w}_4, \mathbf{x}_4 \rangle = -0.0165$ and plot the empirical rejection probability in Figure 2 (c) and (d). Clearly the empirical rejection probability does not shrink towards zero, thus both the angle and the inner product are not preserved.

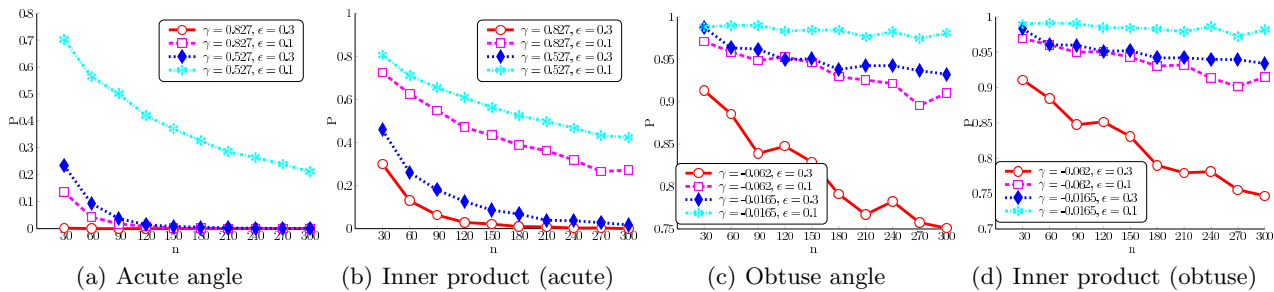

(a) Acute angle     (b) Inner product (acute)     (c) Obtuse angle     (d) Inner product (obtuse)

*Figure 2.* The empirical rejection probability plot for (a) acute angles, (b) inner product with an acute angle, (c) obtuse angles and (d) inner product with an obtuse angle. It demonstrates that angle is preserved better than the inner product under random projection when the angle is acute (*i.e.* $\gamma > 0$). The empirical rejection probability for angles is much smaller than that for the inner product under the same conditions. Neither angle (c) nor inner product (d) is preserved when the angle is obtuse (*i.e.* $\gamma < 0$).

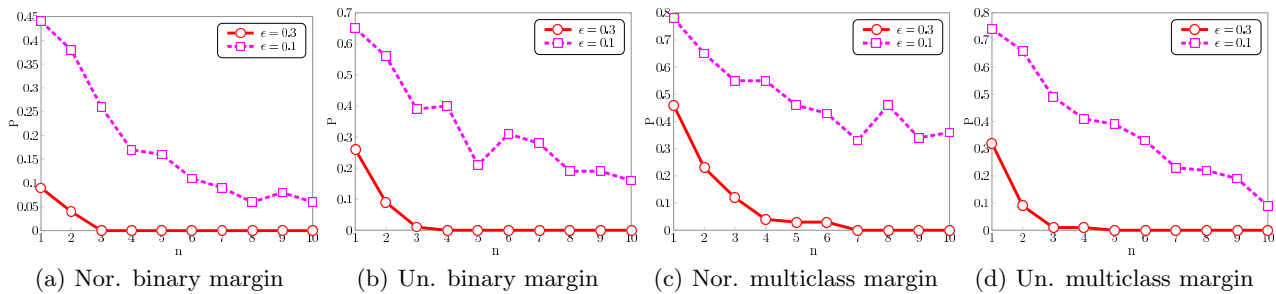

(a) Nor. binary margin     (b) Un. binary margin     (c) Nor. multiclass margin     (d) Un. multiclass margin

*Figure 3.* The empirical rejection probability plots for (a) normalised binary margin, (b) unnormalised binary margin, (c) normalised multiclass margin, (d) unnormalised multiclass margin.

## 5.2. Margin preservation

**Margins** We generated $L$ parallel hyperplanes, where the $L$ is the number of classes. Each class consists of 5 data points $\mathbf{x} \in \mathbb{R}^{100}$ from a hyperplane. We then generated 100 random Gaussian matrices. We used the random matrices to project the data, and then computed both the normalised margin and unnormalised margin. The empirical rejection probability

$$P = 1 - \Pr\left((1 - \epsilon) \leq \frac{\gamma'}{\gamma} < (1 + \epsilon)\right),$$

where $\gamma'$ is the new margin and $\gamma$ is the original margin, was also computed. We show the plots for both binary and multiclass ($L = 3$) cases. As we can see in Figure 3, that the empirical rejection probability decreases ( *i.e.* margins are preserved with higher probability) as the number of projection $n$ increases.

**Implications for SVMs** As has been shown, the results above can be applied even in the case where the data are linearly inseparable, as is often the case in real classification problems. Testing of this method shows that it exhibits a smaller testing error on in the TiCC handwritten digit dataset (van der Maaten,

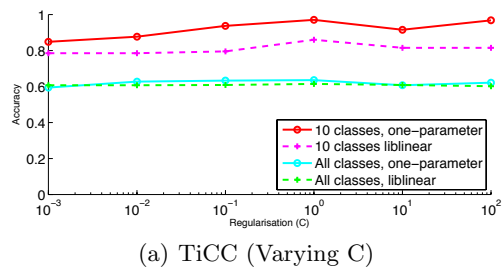

(a) TiCC (Varying C)

*Figure 4.* Higher test accuracy with one-parameter method in TiCC dataset.

2009), for example, than the multiclass SVM (Crammer & Singer, 2001) algorithm in liblinear (Fan et al., 2008). We conjecture that this is due to the significantly reduced dimensionality specifically as a result of the application of the one-parameter method. Projecting features to a lower dimensional space can significantly reduce the model capacity such as VC dimension (Vapnik, 1995). Thus the consequent generalisation bounds can be reduced if the margin preservation is good.

## 6. Conclusion

We have provided an analysis of margin distortion under random projections, described the conditions under which margins are preserved, and given bounds on the margin distortion. We have shown particularly that margin preservation is closely related to acute angle (cosine) preservation and inner product preservation. In doing so we saw that the smaller acute angle, the better the preservation of the angle and the inner product. When the angle is well preserved, the margin is well preserved too. Because of this, the normalised margin is more informative than the unnormalised margin. We have also provided a theoretical underpinning for classification methods which use random projection to achieve multiclass classification with a single model parameter vector.

In contrast to previous work in the area (Balcan et al., 2006) we have shown that it is possible to provide bounds on error free margin preservation without requiring an infinite number of projections, and have done so for arbitrary tolerances, rather than only for half of the original margin. In addition, all of the above has been achieved for multiclass rather than solely binary classifiers. It is worth pointing out that our error free margin is defined on a dataset as traditional margin concepts whereas Balcan et al. (2006)'s error allowed margin is defined on a data distribution.

Though we only showed results for random Gaussian matrix, similar bounds can be achieved for sub-Gaussian distribution as long as a tail bound similar to Lemma 10 holds (see Achlioptas, 2003).

The bounds derived above are conservative, however, as they are based on the union bound over all of the data. The margin is primarily determined by the data on the boundary, however. Even a small distortion of data near the boundary may change the margin significantly, whereas distortion of the data far from the boundary is far less likely to do so. It thus seems likely that the margin bound can be further tightened by taking into account the data distribution.

## Acknowledgments

This work is supported by the Australian Research Council DECRA grant DE120101161. We thank Anders Eriksson for discussion on the counter-example, and Maria-Florina Balcan and Avrim Blum for discussion on error-allowed margin.